\documentclass{article}
\usepackage{spconf,amsmath,graphicx}
\usepackage{multirow}
\usepackage{tabularray}
\usepackage{arydshln}
\usepackage{algorithmicx}
\usepackage[ruled,vlined,boxed]{algorithm2e}
\usepackage{arydshln}
\usepackage{xcolor}
\usepackage{booktabs}
\usepackage{url}

\title{Domain-Specific Improvement on Psychotherapy Chatbot Using Assistant}
\twoauthors
  {Cheng Kang\sthanks{Corresponding Author.}, Daniel Novak, Katerina Urbanova\sthanks{Thanks to Research Center Informatics (No. $CZ.02.1.01/0.0/0.0/16\_019/0000765$), Brain Dynamics (No. $CZ.02.01.01/00/22\_008/0004643$) and Student Grant in Czech Technical University in Prague (NO.$SGS22/165/OHK3/3T/13$) for funding. Katerina Urbanova is also with the National Institute of Mental Health in Prague, Czech Republic}}
  	{Czech Technical University in Prague\\
	Prague, Czech Republic}
  {Yuqing Cheng, Yong Hu\sthanks{The work also was supported during the author's internship at The University of Hong Kong. Yuqing Cheng is also with the Shenzhen Mental Health Centre, China.}}
  	{The University of Hong Kong\\
	Hong Kong}
	
\begin{document}
%\ninept
%
\maketitle
\begin{abstract}
Large language models (LLMs) have demonstrated impressive generalization capabilities on specific tasks with human-written instruction data. However, the limited quantity, diversity, and professional expertise of such instruction data raise concerns about the performance of LLMs in psychotherapy tasks when provided with domain-specific instructions. To address this, we firstly propose Domain-Specific Assistant Instructions based on AlexanderStreet therapy, and secondly we use an adaption fine-tuning method and retrieval augmented generation method to improve pre-trained LLMs. Through quantitative evaluation of linguistic quality using automatic and human evaluation, we observe that pre-trained LLMs on Psychotherapy Assistant Instructions outperform state-of-the-art LLMs response baselines. Our Assistant-Instruction approach offers a half-annotation method to align pre-trained LLMs with instructions, and provide pre-trained LLMs more psychotherapy knowledge. 
\end{abstract}

\section{Introduction}
\label{sec:intro}

\begin{figure}[t]
    \begin{center}
        \includegraphics[width=0.48\textwidth]{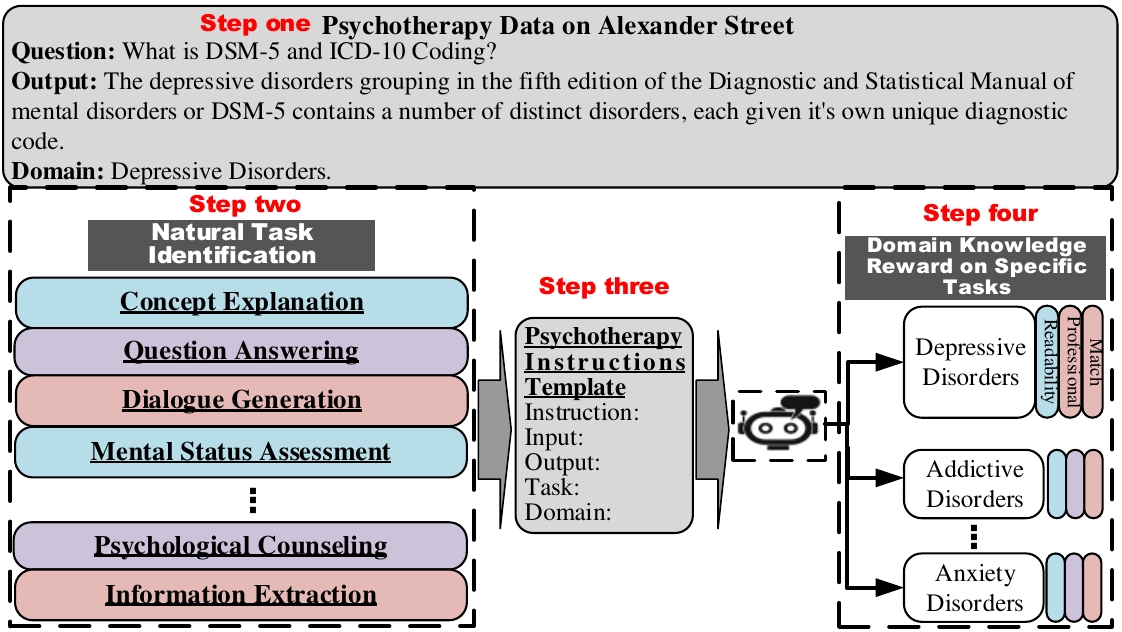}
    \end{center}
    \caption{Schematic representation of Assistant-Instructional prompts in psychotherapy domains. Step one: Task identification; Step two: Knowledge expansion; Step three: Evaluation.}
    \label{Figure4-1}
\end{figure}

Large Language Models (LLMs) have demonstrated impressive generalization capabilities, such as in-context learning \cite{brown2020language}, chain-of-thoughts reasoning \cite{wei2022chain}, and biomedical diagnosing \cite{parmar2022boxbart}. Instruction-tuning of LLMs has enabled them to follow natural language instructions and perform real-world tasks \cite{wang2022super}. Two main methods have been developed for instruction-tuning LLMs: \textbf{(1)} fine-tuning the model on a wide range of tasks using human-annotated prompts and feedback \cite{ouyang2022training}, and \textbf{(2)} supervised fine-tuning using public benchmarks and datasets augmented with manually or automatically generated instructions \cite{wang2022self}. Reinforcement Learning on Human Feedback (RLHF) has proven to be an effective way to improve LLMs in various domains, such as medicine \cite{thirunavukarasu2023large}, knowledge graphs \cite{yang2023chatgpt}, multimodal data fusion \cite{qi2023adaptive} and biomedical applications \cite{yang2023large}, but it comes with a high cost. Self-Instruct tuning \cite{wang2022benchmarking,peng2023instruction} and Guess-Instruction tuning methods have shown better performance in aligning LLMs with human intent by learning from instruction-following data generated by state-of-the-art instruction-tuned teacher LLMs (e.g., GPT-3, GPT-3.5, and even GPT-4). These lines of instruction-tuning research have proven effective in improving the zero and few-shot generalization abilities of LLMs. To improve the professional knowledge of LLMs on psychotherapy domains, our paper presents the psychotherapy Assistant-Instruction approach, which aims to \textbf{(1)} achieve generalization over different psychological consulting tasks and \textbf{(2)} incorporate psychological knowledge into natural common LLMs. Figure \ref{Figure4-1} provides an overview of our proposed approach, in which a single model can perform various NLP tasks in specific psychotherapy domains.

To achieve human-level professional responses in instruction tuning for psychotherapy, we propose a novel approach using GPT-4 as an assistant for Assistant-Instruct tuning (a half self-instruct tuning method) on psychotherapy consulting tasks (Seen in Figure \ref{Figure4-1}). Our method makes the following contributions: \textbf{(a)} it covers a wide range of psychological topics and incorporating feedback knowledge generated by GPT-4. \textbf{(b)} it absorbs psychotherapy knowledge from professional data and enables them to generate content close to GPT-4. \textbf{(c)} it demonstrates the effectiveness of using assistant LLMs-revised instruction data to tune LLMs in psychotherapy domains, providing practical insights to build a general-purpose LLM-following agent powered by assistant LLMs (e.g., GPT-4).

\section{Method}
\label{sec:method}

\subsection{Assistant on Annotation and Task Identification}

To arrange psychotherapy data to correct tasks, such as (1) concept explanation, (2) question answering, (3) mental status assessment, (4) psychological counseling and (5) information extraction, (6) dialogue generation, (7) sentiment analysis, (8) event ordering, we use an assistant LLM -- GPT-4 to identify which task the human-constructed instruction should be. We directly prompt the LLM in a few-shot way to determine this, using 8 classification instructions from the seed tasks. The prompting template is shown in Table 1.

\begin{table}[h]
\label{Table1}
\scriptsize
\centering
\begin{tabular}{ | p{0.94\linewidth} | }
\hline

\textbf{Can the following task be regarded as a question answering task with finite output on {[}***{]} domain?} \\ 
\textbf{Input:} "JEFFREY MISHLOVE Yeah! Well we’re running out of time… time. I supposed the point is that you’ve been successful in… in developing these devices and… and using them in the laboratory?

\textbf{Output:} "STEPHEN LABERGE Yes! That’s right! Well, not just in the laboratory, but we developed devices that… that do have sensors built under the mass that could be used at home, so that… that of course was one of our major goals which used to have make lucid dreaming available in general to people so that they could make better lucid dream." 

\textbf{Result:} Yes
\\  \hline
\end{tabular}
\caption{Prompt used for identifying the type of tasks.}
\end{table}

\subsection{Assistant on Generation, and Evaluation}

\begin{figure}[t]
\begin{center}
%\framebox[4.0in]{$\;$}
\includegraphics[scale=0.4]{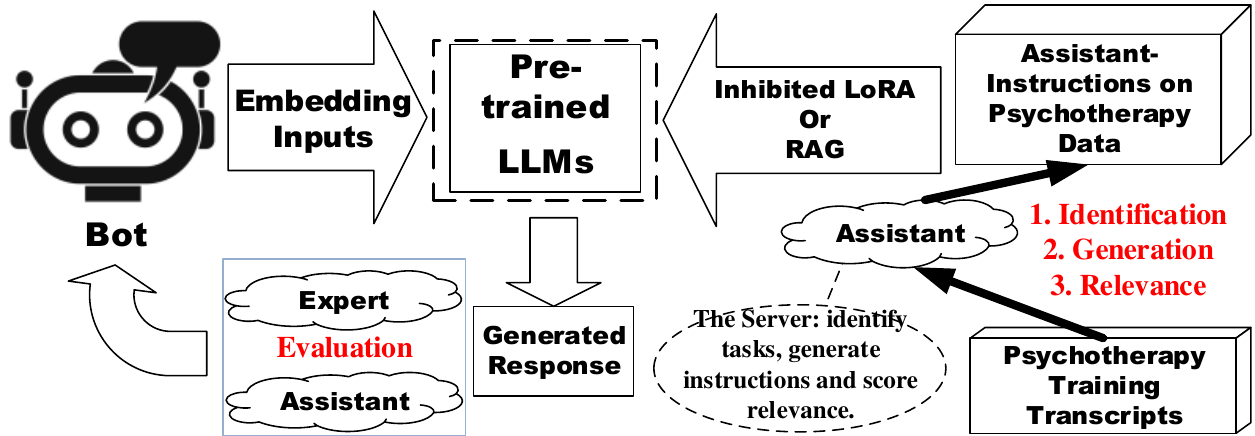}
\end{center}
\caption{The schematic of our system.}
\label{Figure4-2}
\end{figure}

Our approach involves two main steps. Firstly, we optimize formulations that retain the content of the original instructions. We prompt a language model to reformulate the tasks in the core data for each generated task. In some instruction formulations, we embed the input into or add it behind the “INPUT” template -- "We are talking about [***]." -- to emphasize the topic. This manually constructed “INPUT” also captures the content discussed by members of the audience in Alexander Street Video, merging the discussed topic with the point of interest for the audience or visitors. Secondly, following \cite{asai2023self}, we use GPT-4 as an assistant to evaluate the retrieved passage's relevance.The prompting template is shown in Table 2.

\begin{table}[h]
\label{Table2}
\scriptsize
\centering
\begin{tabular}{ | p{0.96\linewidth} | }
\hline

\textbf{Prompt for Generation:} "Make a more professional instruction and output based on given context of conversation in [***] domain. Remove people's names and UNKNOWN. Then, improve them all based on your knowledge. If you cannot do that, output nothing." 

\textbf{Prompt for Evaluation:} "Given an instruction and an output  in [***] domain, rate whether the response appears to be a helpful and
informative answer to the query, from 1 (lowest) - 5 (highest). The detailed criterion is as follows: 5: The response provides a complete, highly detailed, and informative response to the query, fully satisfying the information needs. 4: The response mostly fulfills the need in the query, while there can be some minor improvements such as discussing more detailed information, having better structure of the response, or improving coherence. 3: The response is acceptable, but some major additions or improvements are needed to satisfy users’ needs. 2: The response still addresses the main request, but it is not complete or not relevant to the query. 1: The response is barely on-topic or completely irrelevant.." 
\\  \hline
\end{tabular}
\caption{Prompt used for generation and evaluation.}
\end{table}

\section{Experiments}
\label{sec:experiments}

\subsection{Data Collection}

Alexander Street Press is a website known for its vast collection of video transcripts and recordings from therapy and counseling sessions, covering topics such as depression, abuse, trauma, and mental disorders. The video transcript dataset was specifically collected from the Counseling and Therapy channel on the website. We curated the dataset to include only English-language sessions recorded between 1980 and 2023, resulting in a set of 1,333 videos and accompanying transcripts. After filtering out short-length and non-informative videos, the final dataset comprises 1,179 video transcripts, containing a total of 188,421 dialogue turns. To ensure data quality, we performed a cleaning process to remove Unicode characters, pauses, and other unnecessary elements, resulting in a dataset with 3,141,520 words and a vocabulary size of 30,438.\footnote{\url{https://alexanderstreet.com/}}

On the Alexander Street Press website, most video transcripts and recordings consist of knowledge presentations and counseling talks. For knowledge presentations, there are no instruction questions or instance inputs, and the output is the content presented by the speaker. In the first step, we manually set instructions and instance inputs based on the discussed topics (e.g., Depressive disorders, Addiction, etc.). In the second step, we used the GPT-4 API to revise and generate instructions and instance inputs based on the contents.

\subsection{Models}

We conducted an evaluation of the language models mentioned above for the task of response generation in the psychotherapy domain, specifically focusing on therapeutic counseling. For generating the assistant instructions based on new psychotherapy data, we utilized the GPT-4 API as the Assistant-LLM. To fine-tune the generated instruction data effectively, we employed the inhibition adaption fine-tuning method \cite{kang4551993ina} and self-RAG \cite{asai2023self} on Llama2-7B \cite{touvron2023llama2}, as well as ChatGLM2-6B \cite{du2022glm}. The fine-tuned LLMs were then evaluated by two psychologists on psychotherapy data. The fine-tuning process required two weeks for ChatGLM2-6B/Llama2-7B when using eight NVIDIA Tesla A100 GPUs with 40GB graphic memory cards.\footnote{\url{https://github.com/ChengKang520/psychotherapy-assistant_instruction}}

\subsection{Metrics}

As shown in Table 5, our fine-tuned LLMs are evaluated under two main methods: Automatic evaluation and Human evaluation. For the automatic evaluation, we submit 200 generated answers and use Rouge-L \cite{lin2004rouge} and Fluency \cite{li2022diffusion} to compare the generated output with a reference text and calculate a score for the similarity evaluation. For the human evaluation, we also submit 60 generated outputs (20 examples for each tasks, as we have 3 main tasks) without model and domain information to two professional psychologists, and based on their professional psychotherapy knowledge, they will measure the quality of the generated content according to the questions and answers. There are six grades that they will use to mark the quality of generated content: Extremely  Bad ($1$), Bad ($2$), Neutral ($3$), Acceptable ($4$), Good ($5$) and Very Good ($6$). They should consider the readability, professional and match score (that measures the match with the domain knowledge).

\subsection{Analytic Experiments}

\begin{table}[t]
\label{Table3}
\scriptsize
\centering
\begin{tabular}{p{0.37\linewidth} | p{0.55\linewidth}}
\hline

\textbf{Instruction:}   & \textbf{Instruction:}\\
What suggestions or comments you can provide to address or alleviate the following topics?   & Kindly provide professional suggestions or comments on effectively addressing and alleviating [Depressive Disorders].\\ 
\textbf{Input:}           & \textbf{Input:} \\
We are talking about [Depressive Disorders].    & We are discussing [Depressive Disorders].  \\ 
\textbf{Output:}          & \textbf{Output:}\\
A major depressive episode has a number of characteristic features, which take place most of the day, nearly every day, where relevant. The two main characteristics are a depressed mood, and a noticeably decreased interest in most activities. Other characteristics include trouble sleeping, diminished hunger or pronounced weight loss, trouble concentrating or making decisions, fatigue or loss of energy, recurring suicidal thoughts with or without intent to act on these thoughts, feelings of worthlessness or excessive or inappropriate guilt, psychomotor agitation or retardation, and feelings of low self-worth or pronounced guilt. These symptoms must cause marked distress in the patient's work or social life. For all of these episodes the clinician should ensure that the symptoms are not better explained by another medical condition or by the affects of a substance.    &A major depressive episode is characterized by a range of distinct features that persist for most of the day, nearly every day. The primary indicators include a persistent low mood and a significant decrease in interest or pleasure in most activities. Other symptoms may involve sleep disturbances, changes in appetite or weight, difficulty concentrating or making decisions, fatigue, recurring thoughts of suicide with or without intent to act on them, feelings of worthlessness or excessive guilt, psychomotor agitation or retardation, and feelings of low self-worth. These symptoms should cause marked distress in the individual's work, social life, or daily functioning. When addressing Depressive Disorders, clinicians should ensure that the symptoms are not better explained by another medical condition or the effects of substance use. 
 
 \textcolor{blue}{As professionals, let's explore effective treatment options, such as psychotherapy, cognitive-behavioral therapy, medication, or a combination of these approaches. Encouraging patients to engage in regular physical activity, maintain a healthy diet, and establish a supportive social network can also be beneficial. Furthermore, it is crucial to foster a compassionate and understanding environment to help individuals with Depressive Disorders seek appropriate help and support.}
\\ \hline
\textbf{Perplexity:} 6.71   & \textbf{Perplexity:} 2.15\\ \hline
\end{tabular}
\caption{The manually constructed Instruction and GPT-4 revised Assistant-Instruction on the Depressive Disorder domain.}
\end{table}

\subsubsection{Performance on Revision}

Due to numerous vocabulary and grammar errors in the original psychotherapy data, manual processing and revision of these errors were not feasible. In Table 1, we presented the one command that requires generated instructions from GPT-4. The right panel of Table 3 contains additional information contributed by GPT-4, such as common knowledge about depressive disorders, including psychotherapy, cognitive-behavioral therapy, medication, and a combination of treatment approaches. These recommendations from GPT-4 are considered common-sense by psychologists, but the original psychotherapy data lacked sufficient common knowledge (because it always merely has professional knowledge). The instructions and instance inputs were also refined by GPT-4 based on the given command. Perplexity is a metric that gauges the language model's ability to predict a sequence of words. The perplexity of the right panel is lower than that of the left panel. This revision process makes the content more LLM-friendly, transforming "professional knowledge format" into a more accessible "common knowledge format" (or open-domain knowledge format).

\subsubsection{Generation on Psychotherapy Domain}

When comparing to the content generated by pre-trained ChatGLM2-6B, we observed that the RAG method contains additional professional knowledge (can be seen from Table 4). While this extra knowledge provides more professional information, pre-trained LLMs could not offer further professional insights into autism. When comparing the generated content of ChatGLM2-6B fine-tuned on inhibited LoRA to that of ChatGLM2-6B, we found that the former provides more refinedly revised information (However, this difference is not obvious). For example, it offers insights on professional explanation, such as "\textit{ It is a spectrum disorder, which means that symptoms and severity vary widely among individuals.}".

\begin{table}[t]
\label{Table4}
\scriptsize
\centering
\begin{tabular}{p{0.94\linewidth}}
% \begin{tabular}{@{}cX[c,valign=b]@{}}
\\  \toprule
\textbf{Instruction:}  \\
Please provide professional suggestions or comments to effectively address and alleviate [Autism].\\ 
\textbf{Input:}   \\
We are discussing [Autism].\\ 

\hline
\textbf{Output from Zero-Shot:} \\
Autism is a developmental disorder that can cause significant social, communication and behavioral challenges. \textcolor{blue}{There is no known single cause for autism, but increased awareness and funding support research into possible causes and better ways to treat the disorder}.
\\  \toprule
\textbf{Output from Inhibited LoRA:} \\
Autism is a developmental disorder that can cause significant social, communication and behavioral challenges. \textcolor{blue}{It is a spectrum disorder, which means that symptoms and severity vary widely among individuals}.
\\  \toprule
\textbf{Output from Inhibited LoRA + RAG:} \\
Autism is a developmental disorder that can cause significant social, communication and behavioral challenges. \textcolor{blue}{It is a spectrum disorder, which means that symptoms can range from mild to severe}. Some people with autism may also have other conditions, such as attention deficit hyperactivity disorder (ADHD), anxiety disorders, or epilepsy. \textcolor{red}{Applied Behavior Analysis (ABA) a widely recognized and evidence-based therapy for Autism. ABA techniques focus on promoting positive behaviors and reducing challenging behaviors. Speech and language therapy can help improve communication skills, such as speech articulation, expressive language, and social communication.}
 \\ \bottomrule
\end{tabular}
\caption{Based on Llama2-7B, we illustrate the performance of Zero-Shot, inhibited LoRA Tuned and RAG methods on Psychotherapy data.}
\end{table}

\subsubsection{Evaluation}

We present a performance summary of different instruction-tuning methods applied to two pre-trained LLMs in Table 5. While the Rouge and Fluency evaluation results show improvement with the use of Assistant-Instruction. To validate the performance, we use a selected portion of psychotherapy data as a validation set. Through content revising and leveraging additional common knowledge from GPT-4, both of these two LLMs show significant enhancement in matching the revised answers. Pre-trained LLMs can provide clients with comments to address psychological problems, but the quality of generated content may not always be fully accepted by psychologists. From Table 5, we observe that psychologists tend to prefer models that have been fine-tuned on psychotherapy data. As most LLMs lack specialization in a specific domain, they often require more domain-specific knowledge to improve their performance in professional domains. Because LLMs have been pre-trained on a vast corpus, giving them an inherent advantage in readability, and the size of tokens used does not seem to affect their performance significantly. Regarding the professionalism of the generated content, the psychologists gave higher scores to models that had been fine-tuned on psychotherapy instruction data compared to the corresponding original LLMs. 

\subsubsection{Human Evaluation Agreement} 
To assess the reliability of our human evaluation, we conducted an inner-rater agreement analysis \cite{wang2022self} between our two evaluators. We used Cohen's $\kappa$ to measure inter-rater agreement for categorical items. The 6-level rating scale (ranging from $1$ to $6$) was treated as a categorical variable for each aspect under consideration. The resulting $\kappa$ value was $0.63$, indicating a moderate level of agreement according to common practice. Furthermore, we computed the Spearman correlation coefficient $\rho$ between the ratings of our two evaluators, treating the ratings as ordinal variables (ranging from $1$ to $6$). The obtained coefficient was $\rho = 0.81$, demonstrating a high correlation between the two evaluators. These results indicate a reasonably reliable human evaluation process for our study.

% \usepackage{multirow}
\begin{table}[t]
\label{Table5}
\scriptsize
\centering
\begin{tabular}{cccccc}
\hline
\multicolumn{6}{c}{\textbf{Inhibited LoRA Finetuning (without / with Asisstant-Instruction)}}                                                                                                                       \\ \hline
\multicolumn{1}{c|}{\multirow{2}{*}{\begin{tabular}[c]{@{}c@{}}Pretrained\\ LLM\end{tabular}}} & \multicolumn{2}{c|}{Automatic}       & \multicolumn{3}{c}{Human Evaluation} \\
\multicolumn{1}{c|}{}                                                                          & Rouge-L $\uparrow$ & \multicolumn{1}{c|}{Fluency $\downarrow$} & Read       & Prof       & Match      \\ \hline
\multicolumn{1}{c|}{ChatGLM2-7B}                                                               & 24.3/27.1  & \multicolumn{1}{c|}{49.4/48.7}    & 4.8/4.9           & 2.9/3.3           & 2.1/2.5           \\
\multicolumn{1}{c|}{Llama2-7B}                                                                 & 15.1/16.9  & \multicolumn{1}{c|}{20.9/20.5}    & 5.0/5.2           & 3.0/3.2           & 1.9/2.3           \\ \hline
\multicolumn{6}{c}{\textbf{Retravel Augmented Generation (without / with Asisstant-Instruction)}}                                                                                                                   \\ \hline
\multicolumn{1}{c|}{\multirow{2}{*}{\begin{tabular}[c]{@{}c@{}}Pretrained\\ LLM\end{tabular}}} & \multicolumn{2}{c|}{Automatic}       & \multicolumn{3}{c}{Human Evaluation} \\
\multicolumn{1}{c|}{}                                                                          & Rouge-L $\uparrow$ & \multicolumn{1}{c|}{Fluency $\downarrow$} & Read       & Prof       & Match      \\ \hline
\multicolumn{1}{c|}{ChatGLM2-7B}                                                               & 25.1/32.8  & \multicolumn{1}{c|}{56.4/46.7}    & 4.6/5.3           & 3.9/4.2           & 2.9/3.3           \\
\multicolumn{1}{c|}{Llama2-7B}                                                                 & 15.4/22.4  & \multicolumn{1}{c|}{30.3/20.7}    & 4.8/5.2           & 3.7/4.1           & 3.0/3.4           \\ \hline
\end{tabular}
\caption{ For evaluating the performance of LLMs on psychotherapy domain, two methods - inhibited LoRA and RAG - were used on two pre-trained LLMs have been tuned on Assistant-Instruction using .}
 % ROUGH-L and LMentry are used to match the content with the lable. Psychologist Evaluation (Psych-Eval) is a human-based evaluation method that considered the readability, professional and match of generated content.
\end{table}

%\vfill
%\pagebreak

\section{Conclusion}
\label{conclusion}

We propose a novel method called ASSISTANT-INSTRUCT for fine-tuning or retrieving information from language models (LMs) to improve their instruction-following ability. This method combines both common knowledge and psychotherapy professional knowledge to generate instruction data with the help of experts. It retains the general knowledge already present in pre-trained LMs and incorporates psychotherapy-specific knowledge from expert-presented instructions. To enhance fine-tuning, as well as retrieval knowledge, we format the psychotherapy data, such as presentations, talks, and conversations, to make it more compatible with LMs. Human evaluation of this method demonstrates significant improvement compared to existing instruction methods. ASSISTANT-INSTRUCT can serve as an initial step to align pre-trained LMs with LM-revised instructions, and further research can build upon this method to enhance instruction-following models.

%\section{Acknowledgment}
%\label{acknowledgment}

\bibliographystyle{IEEEbib}
\bibliography{refs}

\begin{thebibliography}{10}

\bibitem{brown2020language}
Tom Brown, Benjamin Mann, Nick Ryder, Melanie Subbiah, Jared~D Kaplan, Prafulla
  Dhariwal, Arvind Neelakantan, Pranav Shyam, Girish Sastry, Amanda Askell,
  et~al.,
\newblock ``Language models are few-shot learners,''
\newblock {\em Advances in neural information processing systems}, vol. 33, pp.
  1877--1901, 2020.

\bibitem{wei2022chain}
Jason Wei, Xuezhi Wang, Dale Schuurmans, Maarten Bosma, Fei Xia, Ed~Chi, Quoc~V
  Le, Denny Zhou, et~al.,
\newblock ``Chain-of-thought prompting elicits reasoning in large language
  models,''
\newblock {\em Advances in Neural Information Processing Systems}, vol. 35, pp.
  24824--24837, 2022.

\bibitem{parmar2022boxbart}
Mihir Parmar, Swaroop Mishra, Mirali Purohit, Man Luo, M~Hassan Murad, and
  Chitta Baral,
\newblock ``In-boxbart: Get instructions into biomedical multi-task learning,''
\newblock {\em arXiv preprint arXiv:2204.07600}, 2022.

\bibitem{wang2022super}
Yizhong Wang, Swaroop Mishra, Pegah Alipoormolabashi, Yeganeh Kordi, Amirreza
  Mirzaei, Anjana Arunkumar, Arjun Ashok, Arut~Selvan Dhanasekaran, Atharva
  Naik, David Stap, et~al.,
\newblock ``Super-naturalinstructions: Generalization via declarative
  instructions on 1600+ nlp tasks,''
\newblock {\em arXiv preprint arXiv:2204.07705}, 2022.

\bibitem{ouyang2022training}
Long Ouyang, Jeffrey Wu, Xu~Jiang, Diogo Almeida, Carroll Wainwright, Pamela
  Mishkin, Chong Zhang, Sandhini Agarwal, Katarina Slama, Alex Ray, et~al.,
\newblock ``Training language models to follow instructions with human
  feedback,''
\newblock {\em Advances in Neural Information Processing Systems}, vol. 35, pp.
  27730--27744, 2022.

\bibitem{wang2022self}
Yizhong Wang, Yeganeh Kordi, Swaroop Mishra, Alisa Liu, Noah~A Smith, Daniel
  Khashabi, and Hannaneh Hajishirzi,
\newblock ``Self-instruct: Aligning language model with self generated
  instructions,''
\newblock {\em arXiv preprint arXiv:2212.10560}, 2022.

\bibitem{thirunavukarasu2023large}
Arun~James Thirunavukarasu, Darren Shu~Jeng Ting, Kabilan Elangovan, Laura
  Gutierrez, Ting~Fang Tan, and Daniel Shu~Wei Ting,
\newblock ``Large language models in medicine,''
\newblock {\em Nature Medicine}, pp. 1--11, 2023.

\bibitem{yang2023chatgpt}
Linyao Yang, Hongyang Chen, Zhao Li, Xiao Ding, and Xindong Wu,
\newblock ``Chatgpt is not enough: Enhancing large language models with
  knowledge graphs for fact-aware language modeling,''
\newblock {\em arXiv preprint arXiv:2306.11489}, 2023.

\bibitem{qi2023adaptive}
Wen Qi, Haoyu Fan, Hamid~Reza Karimi, and Hang Su,
\newblock ``An adaptive reinforcement learning-based multimodal data fusion
  framework for human--robot confrontation gaming,''
\newblock {\em Neural Networks}, vol. 164, pp. 489--496, 2023.

\bibitem{yang2023large}
Rui Yang, Ting~Fang Tan, Wei Lu, Arun~James Thirunavukarasu, Daniel Shu~Wei
  Ting, and Nan Liu,
\newblock ``Large language models in health care: Development, applications,
  and challenges,''
\newblock {\em Health Care Science}, 2023.

\bibitem{wang2022benchmarking}
Yizhong Wang, Swaroop Mishra, Pegah Alipoormolabashi, Yeganeh Kordi, Amirreza
  Mirzaei, Anjana Arunkumar, Arjun Ashok, Arut Selvan~Dhanasekaran, Atharva
  Naik, David Stap, et~al.,
\newblock ``Benchmarking generalization via in-context instructions on 1,600+
  language tasks,''
\newblock {\em arXiv e-prints}, pp. arXiv--2204, 2022.

\bibitem{peng2023instruction}
Baolin Peng, Chunyuan Li, Pengcheng He, Michel Galley, and Jianfeng Gao,
\newblock ``Instruction tuning with gpt-4,''
\newblock {\em arXiv preprint arXiv:2304.03277}, 2023.

\bibitem{asai2023self}
Akari Asai, Zeqiu Wu, Yizhong Wang, Avirup Sil, and Hannaneh Hajishirzi,
\newblock ``Self-rag: Learning to retrieve, generate, and critique through
  self-reflection,''
\newblock {\em arXiv preprint arXiv:2310.11511}, 2023.

\bibitem{kang4551993ina}
Cheng Kang, Jindich Prokop, Lei Tong, Huiyu Zhou, Yong Hu, and Daniel Novak,
\newblock ``Ina: Inhibition adaption on pre-trained language models,''
\newblock {\em Available at SSRN 4551993}.

\bibitem{touvron2023llama2}
Hugo Touvron, Louis Martin, Kevin Stone, Peter Albert, Amjad Almahairi, Yasmine
  Babaei, Nikolay Bashlykov, Soumya Batra, Prajjwal Bhargava, Shruti Bhosale,
  et~al.,
\newblock ``Llama 2: Open foundation and fine-tuned chat models,''
\newblock {\em arXiv preprint arXiv:2307.09288}, 2023b.

\bibitem{du2022glm}
Zhengxiao Du, Yujie Qian, Xiao Liu, Ming Ding, Jiezhong Qiu, Zhilin Yang, and
  Jie Tang,
\newblock ``Glm: General language model pretraining with autoregressive blank
  infilling,''
\newblock in {\em Proceedings of the 60th Annual Meeting of the Association for
  Computational Linguistics (Volume 1: Long Papers)}, 2022, pp. 320--335.

\bibitem{lin2004rouge}
Chin-Yew Lin,
\newblock ``Rouge: A package for automatic evaluation of summaries,''
\newblock in {\em Text summarization branches out}, 2004, pp. 74--81.

\bibitem{li2022diffusion}
Xiang Li, John Thickstun, Ishaan Gulrajani, Percy~S Liang, and Tatsunori~B
  Hashimoto,
\newblock ``Diffusion-lm improves controllable text generation,''
\newblock {\em Advances in Neural Information Processing Systems}, vol. 35, pp.
  4328--4343, 2022.

\end{thebibliography}

\end{document}